\documentclass[11pt,a4paper]{article}
\usepackage[hyperref]{acl2020}
\usepackage{times}
\usepackage{latexsym}
\usepackage{booktabs}
\usepackage{float}

\usepackage{microtype}

\usepackage{amsfonts}
\usepackage{bm}
\usepackage{subfiles}
\usepackage[utf8]{inputenc}
\usepackage[inline]{enumitem}
\usepackage{graphicx}
\usepackage{caption}
\usepackage{subcaption}

\newcommand*\circled[1]{\raisebox{.5pt}{\textcircled{\raisebox{-.9pt} {#1}}}}

\usepackage[disable]{todonotes}

\aclfinalcopy
\def\aclpaperid{58}

\newcommand{\dobib}{
    \bibliography{themes}
    \bibliographystyle{acl_natbib}
}

\graphicspath{{figures/}{../figures/}}

\newcommand{\nstm}{NSTM}

\makeatletter
\newcommand{\authorfootnote}[1]{{\normalfont \textsuperscript{\@fnsymbol{#1}}}}
\makeatother

\renewcommand*{\thefootnote}{\fnsymbol{footnote}}

\title{\nstm: Real-Time Query-Driven News Overview Composition at Bloomberg}

\newcommand*\bb[0]{{\normalfont \textsuperscript{1}}}
\newcommand*\on[0]{{\normalfont \textsuperscript{2}}}

  \author{
  Joshua Bambrick\bb, Minjie Xu\bb, Andy Almonte\bb, Igor Malioutov\bb,\\
  \textbf{Guim Perarnau\bb, Vittorio Selo\bb, Iat Chong Chan\on\textsuperscript{, }\authorfootnote{1}} \\
  \bb{}Bloomberg, London, United Kingdom\\ \on{}OakNorth, London, United Kingdom \\
  \small{\bb\texttt{\{jbambrick7,mxu161,aalmonte2,imalioutov,gperarnau,vselo\}@bloomberg.net}} \\
  \small{\on\texttt{iat.chan@oaknorth.com}}
  }

\date{}

\begin{document}
\renewcommand{\dobib}{}
\maketitle

\footnotetext[1]{Order reflects writing contributions; M.X., I.C.C., and J.B. designed and developed a prototype of the system; All implemented the production system; A.A. managed the project. I.C.C. worked on the project while employed by Bloomberg.}

\renewcommand*{\thefootnote}{\arabic{footnote}}
\setcounter{footnote}{0}

\begin{abstract}

Millions of news articles from hundreds of thousands of sources around the globe appear in news aggregators every day. Consuming such a volume of news presents an almost insurmountable challenge. For example, a reader searching on Bloomberg's system for news about the U.K. would find 10,000 articles on a typical day. Apple Inc., the world's most journalistically covered company, garners around 1,800 news articles a day.

We realized that a new kind of summarization engine was needed, one that would condense large volumes of news into short, easy to absorb points. The system would filter out noise and duplicates to identify and summarize key news about companies, countries or markets.

When given a user query, Bloomberg's solution, Key \underline{N}ew\underline{s} \underline{T}he\underline{m}es (or \nstm{}), leverages state-of-the-art semantic clustering techniques and novel summarization methods to produce comprehensive, yet concise, digests to dramatically simplify the news consumption process.

\nstm{} is available to hundreds of thousands of readers around the world and serves thousands of requests daily with sub-second latency. At ACL 2020, we will present a demo of \nstm{}.

\end{abstract}

\section{Introduction}
\label{sec:intro}

In many domains, finding contextually-important news as fast as possible is a key goal. With millions of articles published around the globe each day, quickly finding relevant and actionable news can mean the difference between success and failure.

When provided with a search query, a traditional system returns links to articles sorted by relevance. However, users typically encounter (near) duplicate or overlapping articles, making it hard to quickly identify key events and easy to miss less-reported stories. Moreover, news headlines are frequently sensational, opaque, or verbose, forcing readers to open and read individual articles.

For illustration, imagine an analyst sees the price of Amazon.com stock drop and wants to know why. With a traditional system, they would search for news on the company and wade through many stories ($307$ in this case\footnote{The corresponding overview can be found in Appendix~\ref{appendix:nstm-screenshots}.}), often with duplicate information or unhelpful headlines, to slowly build up a full picture of what the key events were.

\begin{figure*}
    \centering
    \includegraphics[width=16cm]{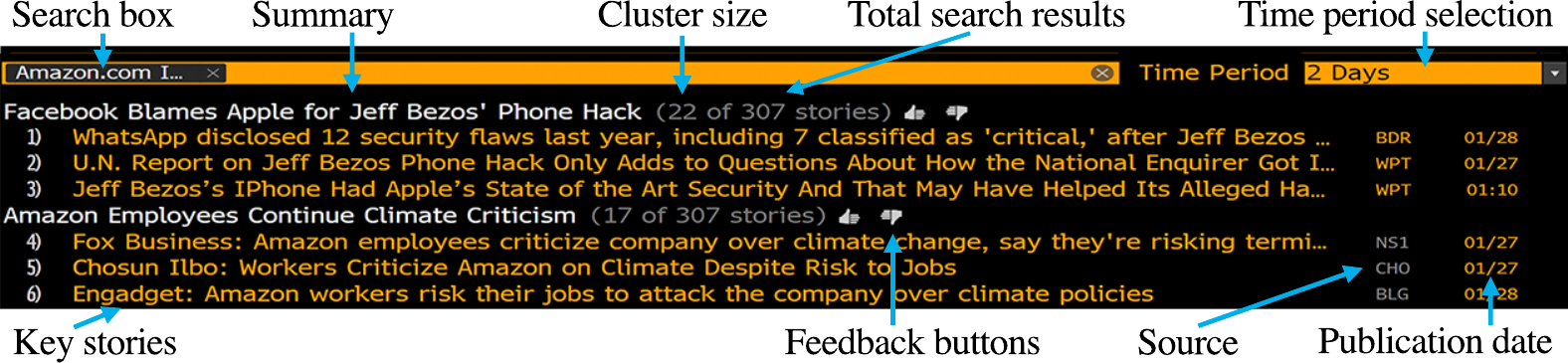}
    \caption{A query-based UI for \nstm{} showing two themes. The un-cropped screenshot is in Appendix~\ref{appendix:nstm-screenshots}.}
    \label{fig:2-themes}
\end{figure*}

By contrast, using \emph{\nstm{} (Key News Themes)}, this same analyst can search for `Amazon.com', over a given time horizon, and promptly receive a concise and comprehensive overview of the news, as shown in Fig.~\ref{fig:2-themes}. We tackle the challenges involved with consuming vast quantities of news by leveraging modern techniques to semantically cluster stories, as well as innovative summarization methods to extract succinct, informational summaries for each cluster. A handful of \emph{key stories} are then selected from each cluster. We define a (story cluster, summary, key stories) triple as one \emph{theme} and an ordered list of themes as an \emph{overview}.

\nstm{} works at web scale but responds to arbitrary user queries with sub-second latency. It is deployed to hundreds of thousands of users around the globe and serves thousands of requests per day.

\section{Design Goals}
\label{sec:requirements}
We focus on the scenario where a news search query can render many matching news articles, from tens up to hundreds of thousands. The task is to create a succinct overview of the results to help our users to easily grasp the gist of them without combing through the individual articles.

Since the matching articles often cover various aspects and events, \nstm{} must first cluster related stories to form a clear separation among them.

Furthermore, the system must extract a concise (up to $50$ characters, or roughly 6 tokens) summary for each cluster. It needs to be short enough to be understandable to humans with a single glance, but also rich enough to retain critical details from a minimal `who-does-what' stub, so the most popular noun phrase or entity alone will not suffice. Such conciseness also helps when screen space is limited (for context-driven applications or mobile devices).

From each cluster, \nstm{} must surface a few key stories to provide a sample of its contents. The clusters themselves should also be ranked to highlight the most important few in limited screen space.

Finally, the system must be fast. It may only take up to a few seconds for the slowest queries.

\paragraph{Main technical challenges:}
\textbf{1)} There is no public dataset corresponding to this overview composition problem with all the requirements set above, so we were required to either define new (sub-)tasks and collect new annotations, or select techniques by intuition, implement them, and iterate on feedback; \textbf{2)} Generating summaries which are simultaneously accurate, informational, fluent, and highly concise necessitates careful and innovative choices of summarization techniques; \textbf{3)} Supporting arbitrary user searches in real-time places significant performance requirements on the system whilst also setting a high bar for its robustness.

\section{Related Work}
\label{sec:related-work}

A comparable system is Google News' `Full Coverage' feature\footnote{https://www.blog.google/products/news/new-google-news-ai-meets-human-intelligence/}, which groups stories from different sources, akin to our clustering approach. However, it doesn't offer summarization and its clustered view is unavailable for arbitrary search queries.

SUMMA~\citep{liepins-etal-2017-summa} is another comparable system which integrates a variety of NLP components and provides support for numerous media and languages, to simultaneously monitor several media broadcasts. SUMMA applies the online clustering algorithm by \citet{aggarwal2006framework} and the extractive summarization algorithm by \citet{almeida-martins-2013-fast}. In contrast to \nstm{}, SUMMA focuses on scenarios with continuous multimedia and multilingual data streams and produces much longer summaries.

\section{Approach}

\subsection{Architecture}

\begin{figure*}
    \centering
    \includegraphics[width=16cm]{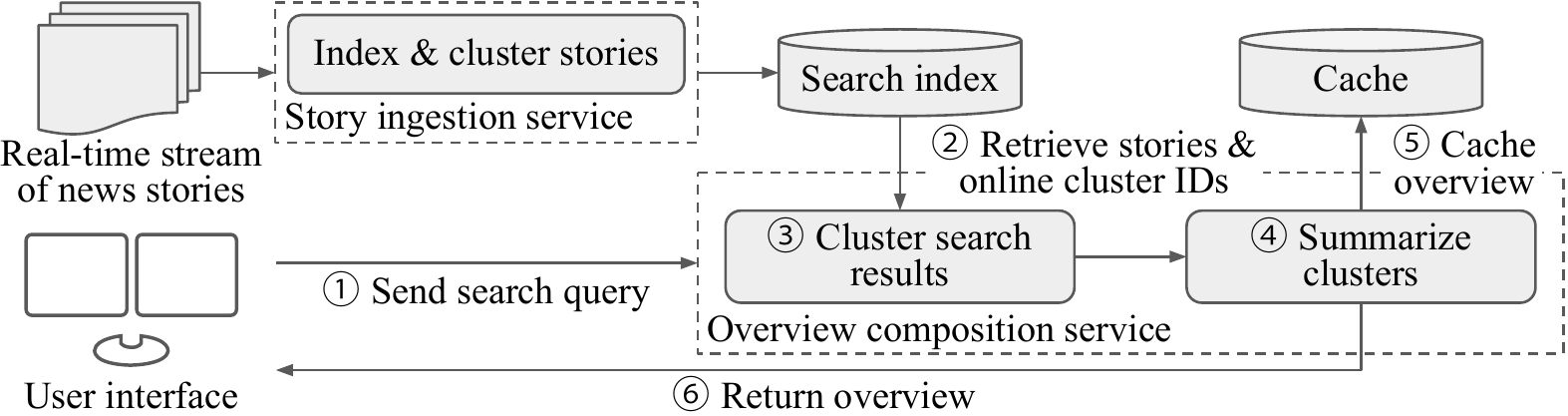}
    \caption{The architecture of \nstm{}. The digits indicate the order of execution whenever a new request is made.}
    \label{fig:architecture}
\end{figure*}

The functionality of NSTM can be formulated as: given a search query, generate a ranked list (\emph{overview}) of the key \emph{themes}, or (news cluster, summary, key stories) triples, that concisely represent the most important matching news events.

Fig.~\ref{fig:architecture} depicts the system's architecture. The \emph{story ingestion service} processes millions of published news stories each day, stores them in a search index, and applies online clustering to them. When a search query is submitted via a user interface (\circled{1} in the diagram), the \emph{overview composition service} retrieves matching stories and their associated online cluster IDs from the search index (\circled{2}). The system then further clusters the retrieved online clusters into the final clusters, each corresponding to one theme (\circled{3}). For each such cluster, the system extracts a concise summary and a handful of key stories to reflect the cluster's contents (\circled{4}). This creates a set of themes, which \nstm{} ranks to create the final overview. Lastly, the system caches the overview for a limited time to support future reuse (\circled{5}) before returning it to the UI (\circled{6}).

\subsection{News Search}
\label{sec:search}
The first step in the \nstm{} pipeline is to retrieve relevant news stories (\circled{1} in Fig.~\ref{fig:architecture}), for which we leverage a customized in-house news search engine based on Apache Solr.\footnote{http://lucene.apache.org/solr/} This supports searches based on keywords, metadata (such as news source and time of ingestion), and tags generated during ingestion (such as topics, regions, securities, and people). For example, \texttt{TOPIC:ECOM AND NOT COMPANY:AMZN}\footnote{This is Bloomberg's internal news search query syntax, which maps closely to the final query submitted to Solr.} will retrieve all news about `E-commerce' but exclude Amazon.com.

\nstm{} uses Solr's \emph{facet} functionality to surface the largest $k$ online clusters (detailed in Sec.~\ref{sec:clustering}) in the search results, before returning $n$ stories from each. This tiered approach offers better coverage and scalability than direct story retrieval.

\subsection{Clustering}
\subsubsection{News Embedding and Similarity}
\label{sec:embedding}
At the core of any clustering system is a similarity metric. In \nstm{}, we define the similarity between two articles as the cosine similarity between their embeddings as computed by NVDM~\citep{nvdm}, i.e., $\tau(d_1,d_2)=0.5(\cos(\bm{z}_1,\bm{z}_2)+1)$, where $\bm{z}\in\mathbb{R}^n$ denotes the NVDM embedding.

Our choice is motivated by two observations: \textbf{1)} The generative model of NVDM is based on bag-of-words (BoW) and $P(w|\bm{z})=\sigma(W^\top\bm{z})$ where $\sigma$ is the softmax function, $W\in\mathbb{R}^{n\times V}$ is the word embedding matrix in the decoder and $V$ is the size of the vocabulary. This resembles the latent topic structure popularized by LDA~\citep{lda} which has proven effective in capturing textual semantics. Additionally, the use of cosine similarities is naturally motivated by the fact that the generative model is directly defined by the dot-product between the story embedding ($z$) and a shared vocabulary embedding ($W$). \textbf{2)} NVDM's Variational Autoencoder (VAE)~\citep{KingmaW13VAE, RezendeMW14VAE} framework makes the inference procedure much simpler than LDA and it also supports decoder customizations. For example, it allows us to easily integrate the idea of introducing a learnable common background word distribution into the generative model~\cite{sif}.

We trained the model on an internal corpus of $1.85$M news articles, using a vocabulary of size about $200$k and a latent dimension $n$ of $128$.

\dobib

\subsubsection{Clustering Stages}
\label{sec:clustering}
We divide clustering into two stages in the pipeline, \textbf{1)} online incremental clustering at story ingestion time, and \textbf{2)} hierarchical agglomerative clustering (HAC) at query time (\circled{3} in Fig.~\ref{fig:architecture}). The former is used to produce query-agnostic \emph{online clusters} at a relatively low cost to handle the daily influx of millions of news stories. These clusters reduce the computational cost at query time. However, due to its online nature, over-fragmentation, among other quality issues, occurs in the resulting clusters. This necessitates further refinement at query time when an offline HAC step is performed on top of the retrieved online clusters. A similar, but more complicated, design was adopted in \citet{yahoo} for clustering real-time news search results.

At both stages, we compute the cluster embedding $\bm{z}_c\in\mathbb{R}^n$ as the mean of all the story embeddings therein, and evaluate similarities between clusters (individual stories are taken as singleton clusters) using the metric $\tau$ defined in Sec.~\ref{sec:embedding}.

For online clustering, we apply an in-house implementation which uses a distributed pool of workers to reduce latency and increase throughput. It merges each incoming story with the closest cluster if the similarity is within a parameterized threshold and otherwise creates a new singleton cluster.

For HAC, we apply \texttt{fastcluster}\footnote{https://www.jstatsoft.org/article/view/v053i09}~\citep{fastcluster} to construct the dendrogram. We use complete linkage to encourage more congruent clusters and then form flat clusters by cutting the dendrogram at the same (height) threshold. To further reduce fragmentation where similar clusters are left un-clustered, we apply HAC twice recursively.

To find a reasonable similarity threshold, we manually annotated just over 1k pairs of news articles. Each annotator indicated whether they would expect to see the articles grouped together or not in an overview. We then selected the threshold which achieved the highest F\textsubscript{1} score on this binary classification task, which was $0.86$.

\dobib

\subsection{Summary Extraction}
\label{sec:summary}
Clustering search results~\citep{yahoo} is a meaningful step towards creating a useful overview. With NSTM, we push this one step further by additionally generating a concise, yet still human-readable, summary for each cluster (\circled{4} in Fig.~\ref{fig:architecture}).

\begin{figure*}
    \centering
    \includegraphics[width=16cm]{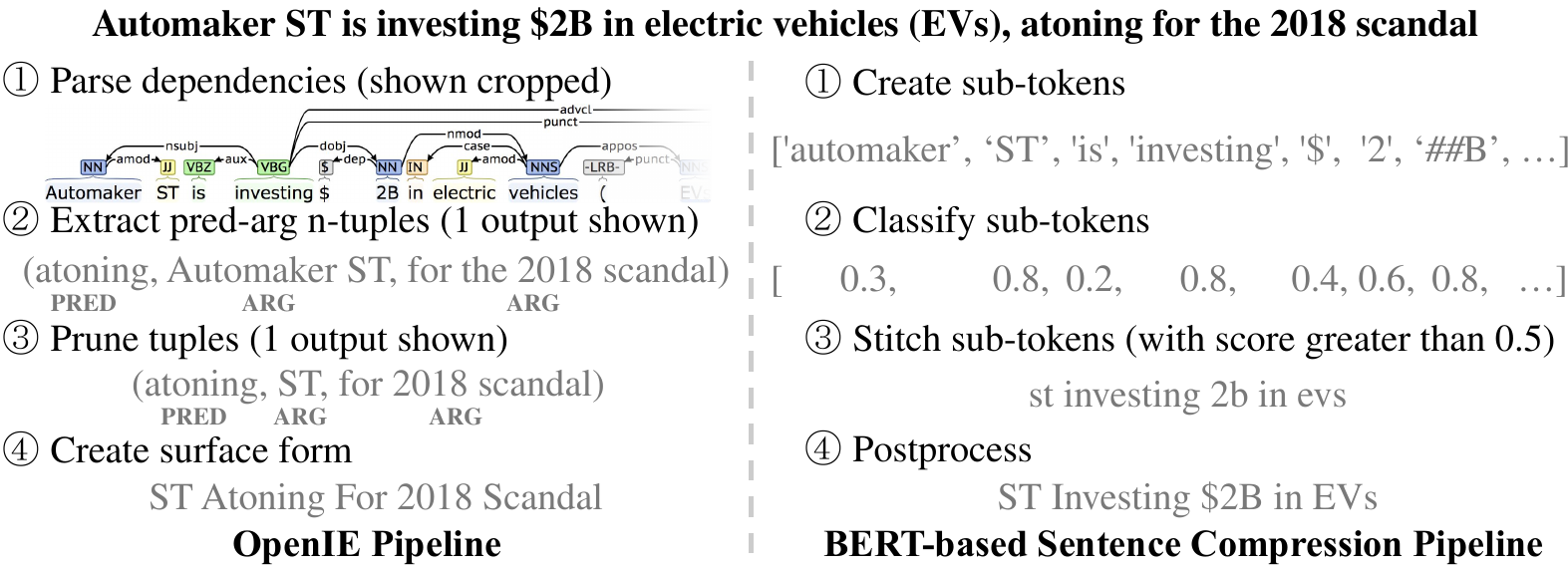}
    \caption{Illustrations of the symbolic OpenIE (left) and neural sentence compression (right) candidate extraction pipelines. We apply both, to render a diverse pool of candidate summaries, and use a ranker to select the best.}
    \label{fig:summary-extraction}
\end{figure*}

Due to the unique style of the summary explained in Sec.~\ref{sec:requirements}, the scarcity of training data makes it hard to train an end-to-end seq2seq~\citep{SutskeverVL14seq2seq} model, as is typical for abstractive summarization. Also, this technique would only offer limited control over the output. Hence, we opt for an extractive method, leveraging OpenIE~\citep{banko2007open} and a BERT-based~\citep{devlin-etal-2019-bert} sentence compressor (both illustrated in Fig.~\ref{fig:summary-extraction}) to surface a pool of sub-sentence-level candidate summaries from the headline and the body, which are then scored by a ranker.

\subsubsection{OpenIE-based Tuple Extraction}
\label{sec:openie}

Open Domain Information Extraction (OpenIE) presents an unsupervised approach to extract summary candidates from an input sentence.

First, we construct a dependency parse tree of the sentence, using a model based on \citet{kiperwasser2016simple} (\circled{1} in Fig.~\ref{fig:summary-extraction}).

From this tree, we extract predicate-argument $n$-tuples using an adapted reimplementation of PredPatt~\citep{white-EtAl:2016:EMNLP2016} (\circled{2}). The tuples represent nested proto-semantic parses of the sentence, and typically correspond to well-formed phrases. This method applies rules cast over Universal Dependencies~\citep{nivre2016universal} so syntactic patterns are unlexicalized and language-neutral.

We then prune these tuples (\circled{3}), applying rules which reduce the arguments to their syntactic heads, while heuristics keep named entities and multi-word expressions intact. We recursively intersect the resulting tuples to create more tuples.

Finally, to render summary candidates, we create a titlecased surface form of each tuple (\circled{4}).

\subsubsection{BERT-based Sentence Compression}
\label{sec:bert-compression}
In addition to the rule-based OpenIE system, we apply a Transfer Learning-based solution, using a novel in-house dataset specific to our sub-task. In particular, we model candidate summary extraction as a `sentence compression' task~\citep{filippova-etal-2015-sentence}, where each story is split into sentences and tokens are classified as \emph{keep} or \emph{delete} to make each sentence shorter, while retaining the key message.

We oversaw the manual annotation of a dataset which maps sentences to compressed equivalents that correspond to summaries. When presented with a news story, annotators selected one sentence and deleted words to create a high quality summary. This rendered $10$k annotations which we randomly partitioned into train ($80\%$) and test ($20\%$) sets.

The task is formulated as sequence tagging, whereby each sub-token (\circled{1} in Fig.~\ref{fig:summary-extraction}), defined using the BERT vocabulary, is classified as \emph{keep} or \emph{delete} (\circled{2}). We implement this using a feedforward layer on top of a Bloomberg-internal pre-trained neural network, akin to the uncased English BERT-Base model, applying an adapated implementation.

To create a compression, we stitch sub-tokens labelled \emph{keep} together (\circled{3}). Lastly, we use postprocessing rules to improve formatting (\circled{4}), such as titlecasing and fixing partial-entity deletion (where only some sub-tokens of a token/entity are deleted).

\dobib

\subsubsection{Summary Candidate Ranking}
\label{sec:candidate-ranking}

Tuple generation and sentence compression provide a pool of summary candidates for individual news stories. These are further aggregated across stories within a cluster to form the final pool. To identify the best summary for the cluster, we trained a sequence-pair model $s_{\bm{\theta}}(a,c)$ to score each candidate $c$ given an article $a$. Such article-level scores for a candidate are computed against all the stories in a cluster and then aggregated (e.g., averaged) to produce the final cluster-level scores, which we use for ranking.

For this purpose, we collected an in-house annotated dataset. We sampled a few thousand news articles and generated $33$k summary candidates from them using OpenIE,\footnote{At this time, we hadn't considered sentence compression.}. Then we asked internal annotators to label each as \emph{Great}, \emph{Acceptable} or \emph{Terrible} were it to be used as a summary for the article, considering both readability and informativeness.

From this dataset, we constructed about $48$k pairwise samples $(c, c')|a$ where $c$ is labelled more favorably than $c'$ for a given common article $a$, and the model $s_{\bm{\theta}}(a,c)$ was then trained to match such preferences using pairwise margin loss, i.e., $\max(0, 1 - s_{\bm{\theta}}(a,c) + s_{\bm{\theta}}(a,c'))$.

We considered a few models, including a parameter-free baseline which scores candidate-article pairs as the dot-product of their NVDM (Sec.~\ref{sec:embedding}) embeddings, i.e., $s=\bm{z}_a^\top\bm{z}_c$. We also considered this model's bilinear extension $s=\bm{z}_a^\top W\bm{z}_c$ where $W$ is the learnable weight matrix. Lastly, we tried neural network models, such as DecAtt~\citep{decatt}. We evaluated these models on a held-out test set with metrics such as pairwise ranking accuracy and NDCG. We opted to productionize the baseline model, since it was the simplest and performed on par with the others.\footnote{E.g., with NDCG\textsubscript{5}, the (untrained) NVDM dot-product yields $0.61$, while the bilinear model and DecAtt yield $0.64$.}

Because NVDM uses a bag-of-words model, this ranker ignores syntax entirely. We believe that its empirical success owes to both the well-formedness of the majority of the candidates and the averaging effect that amplifies the `signal-noise ratio' when the scores are averaged over the cluster.

Empirically, this approach tends to surface `informational' summaries, in contrast to headlines which are often `sensational'. We posit that this is because high-ranked summaries must also be representative of story bodies, not just headlines.

\subsubsection{Combining Summary Candidates}
\label{sec:combining-candidates}

OpenIE and sentence compression offer distinct ways to extract candidates, and we experimented with each as the sole source of summary candidates in our pipeline. On the basis of ROUGE scores~\citep{lin-hovy-2003-automatic, lin-2004-rouge} (details in Appendix~\ref{sec:openie_vs_bert}), the latter provides superior results.

However, in a production system which informs business decisions, we must consider factors which aren't readily captured by metrics which compare generated and `gold' outputs. For example, changing a single word can reverse the meaning of a summary, with only a small change in such scores. Hence, we consider a range of pros and cons.

The sentence compression method is supervised and is trained to produce summaries which can take advantage of news-specific grammatical styles. However, the OpenIE system is much faster and offers greater interpretability and controllability.

Since the neural and symbolic systems provide different advantages, we apply both. This renders a diverse pool of candidate summaries from which the ranker's task is to select the best. At the pooling stage we also impose a length constraint of 50 characters and exclude any longer candidates.

\subsection{Key Story Selection}

As a sample from the full story cluster, \nstm{} selects an ordered list of \emph{key stories} which are deemed to be \emph{representative}. We select these using a heuristic based on intuition and client feedback.

Our approach is to re-cluster all stories in the cluster using HAC (see Sec.~\ref{sec:clustering}), to create a parameterized number of sub-clusters. For each sub-cluster, we select the story that has maximum average similarity $\tau$ (as per Sec.~\ref{sec:embedding}) to the other sub-cluster stories. This strategy is intended to select stories which represent each cluster's diversity.

We sort the key stories by sub-cluster size and time of ingestion, in that order of precedence.

\subsection{Theme Ranking}
We have described how (story cluster, summary, key stories) triples, or \emph{themes}, are created. However, some themes are considered to be more \emph{important} than others since they are more useful to readers. It is tricky to define this concept concretely but we apply proxy metrics in order to estimate an importance \emph{score} for each theme. We rank themes by this score and, in order to save screen space, return only the top few (`key') themes as an \emph{overview}.

The main factor considered in the importance score is the size of the story cluster -- the larger the cluster, the larger the score. This heuristic corresponds to the observation that more important themes tend to be reported on more frequently. Additionally, we consider the entropy of the news sources in the cluster, which corresponds to the observation that more important themes are reported on by a larger number of publishers and reduces the impact of a source publishing duplicate stories.

\subsection{Caching}

Since many user requests are the same or use similar data, caching is useful to minimize response times. When \nstm{} receives a request, it checks whether there is a corresponding overview in the cache, and immediately returns it if so. $99.6\%$ of requests hit the cache and $99\%$ of requests are handled within $215 ms$.\footnote{Computed for all requests over a 90-day period.} In the event of a cache miss, \nstm{} responds in a median time of $723ms$.\footnote{Computed for the top 50 searches over a 7-day period.}

We apply two mechanisms to ensure cache freshness. Firstly, we preemptively invoke \nstm{} using requests that are likely to be queried by users (e.g., most read topics) and re-compose them from scratch at fixed intervals (e.g., every 30 min). Once computed, they are cached. The second mechanism is user-driven: every time a user requests an overview which is not cached, it will be created and added to the cache. The system will subsequently preemptively invoke \nstm{} using this request for a fixed period of time (e.g., 24 hours).

\section{Demonstration}

\nstm{} was deployed to our clients in 2019. Using the UI depicted in Fig.~\ref{fig:2-themes}, users can find overviews for customized queries to help support their work. From this screen, the user can enter a search query using any combination of Boolean logic with tag- or keyword-based terms. They may also alter the period that the overview is calculated over (this UI offers 1 hour, 8 hour, 1 day, and 2 day options).

This interface also allows users to provide feedback via the `thumb' icons or plain-text comments. Of several hundred per-overview feedback submissions, over three quarters have been positive.

\begin{table}
\resizebox{\columnwidth}{!}{
\begin{tabular}{lll}
\toprule   
    {} & Summary & Size \\
    \midrule
    1     & Facebook to Settle Recognition Privacy Lawsuit  & 90 \\
    2     & Facebook Warns Revenue Growth Slowing           & 79 \\
    3     & Facebook Stock Drops 7\% Despite Earnings Beat  & 70 \\
    4     & Facebook to Remove Coronavirus Misinformation   & 49 \\
    5     & Mark Zuckerberg to Launch WhatsApp Payments     & 19 \\
\bottomrule
\end{tabular}
}
\caption{Ranked theme summaries and cluster sizes for `Facebook' (1,176 matching stories) from 31 Jan. 2020.}
\label{table:themes1}
\end{table}

\setlength{\belowcaptionskip}{-0.25pt}
\begin{table}
\resizebox{\columnwidth}{!}{
\begin{tabular}{lll}
\toprule   
    {} & Summary & Size \\
    \midrule
    1     & Britain to Leave the EU                             & 459 \\
    2     & Bank of England Would Keep Interest Rate Unchanged  & 141 \\
    3     & Sturgeon Demands Scottish Independence Vote         & 71  \\
    4     & Pompeo in UK for Trade Talks                        & 45  \\
    5     & Boris Johnson Hails ‘Beginning’ on Brexit Day       & 63  \\
\bottomrule
\end{tabular}
}
\caption{Ranked theme summaries and cluster sizes for `U.K.' (13,858 matching stories) from 31 Jan. 2020.}
\label{table:themes2}
\end{table}

\setlength{\belowcaptionskip}{0pt}

Tables \ref{table:themes1} and \ref{table:themes2} show example theme summaries generated for the queries `Facebook' and `U.K.'. Note that the summaries are quite different from what has previously been studied by the NLP community (in terms of brevity and grammatical style) and that they accurately represent distinct events.

In addition to user-driven settings, \nstm{} can be used to supplement context-driven applications. One example, demonstrated in Appendix~\ref{appendix:tren-screenshots}, uses themes provided by \nstm{} to help explain why companies or topics are `trending'.

\section{Conclusion}

We presented \nstm{}, a novel and production-ready system that composes concise and human-readable news overviews given arbitrary user search queries.

\nstm{} is the first of its kind; it is query-driven, it offers unique news overviews which leverage clustering and succinct summarization, and it has been released to hundreds of thousands of users.

We also demonstrated effective adoption of modern NLP techniques and advances in the design and implementation of the system, which we believe will be of interest to the community.

There are many open questions which we intend to research, such as whether autoregressivity in neural sentence compression can be exploited and how to compose themes over longer time periods.

\bibliography{themes}
\bibliographystyle{acl_natbib}

\clearpage
\appendix

\section{Acknowledgements}
This has been a multi-year project, involving contributions from many people at different stages.

In particular, we thank Miles Osborne, Marco Ponza, Amanda Stent, Mohamed Yahya, Christoph Teichmann, Prabhanjan Kambadur, Umut Topkara, Ted Merz, Sam Brody, and Adrian Benton for reviewing and commenting on the manuscript; We further thank Adela Quinones, Shaun Waters, Mark Dimont, Ted Merz and other colleagues from the News Product group for helping to shape the vision of the system; We also thank Jos\'e Abarca and his team for developing the user interface; We thank Hady Elsahar for helping to improve summary ranking during his internship; Finally, we thank all colleagues (especially those in the Global Data department) who helped to produce high quality in-house annotations and all others who contributed valuable thoughts and time into this work.

\section{End-To-End Evaluation}
\label{sec:openie_vs_bert}

We evaluate the end-to-end \nstm{} system when using the OpenIE (Sec.~\ref{sec:openie}) and the BERT-based sentence compression (Sec.~\ref{sec:bert-compression}) algorithms as the sole source of candidate summaries. We also conducted one experiment where both were used to create a shared pool of candidates (as per Sec.~\ref{sec:combining-candidates}).

We test the system end-to-end using the manually-annotated Single Document Summarization (SDS) test set described in Sec.~\ref{sec:bert-compression}. To implement SDS, our experimental setup assumes that only one story was returned by a search request (as per Sec.~\ref{sec:search}). We evaluate the output from each system with ROUGE~\citep{lin-hovy-2003-automatic, lin-2004-rouge}\footnote{https://github.com/google/seq2seq/blob/master/seq2seq/metrics/rouge.py}. The results are presented in Table~\ref{table:candidate-extraction-results}.

\begin{table}[h]
\resizebox{\columnwidth}{!}{
\begin{tabular}{llll}
\toprule   
    Metric & OpenIE & BSC & Both \\
    \midrule
    ROUGE-1 F\textsubscript{1} & 0.831 & 0.863 & 0.851 \\
    ROUGE-2 F\textsubscript{1} & 0.609 & 0.701 & 0.667 \\
    ROUGE-3 F\textsubscript{1} & 0.530 & 0.640 & 0.599 \\
    ROUGE-4 F\textsubscript{1} & 0.492 & 0.603 & 0.562 \\
    ROUGE-L F\textsubscript{1} & 0.621 & 0.706 & 0.670 \\
\bottomrule
\end{tabular}
}
\caption{ROUGE scores for the Single-Document Summarization task in the end-to-end system, when using OpenIE, BERT-based sentence compression (BSC) and both to construct the pool of candidate summaries.}
\label{table:candidate-extraction-results}
\end{table}

\clearpage

\onecolumn

\section{Screenshots of A Query-Driven User Interface}

\label{appendix:nstm-screenshots}

\begin{figure*}[!ht]
      \centering
      \includegraphics[width=\textwidth]{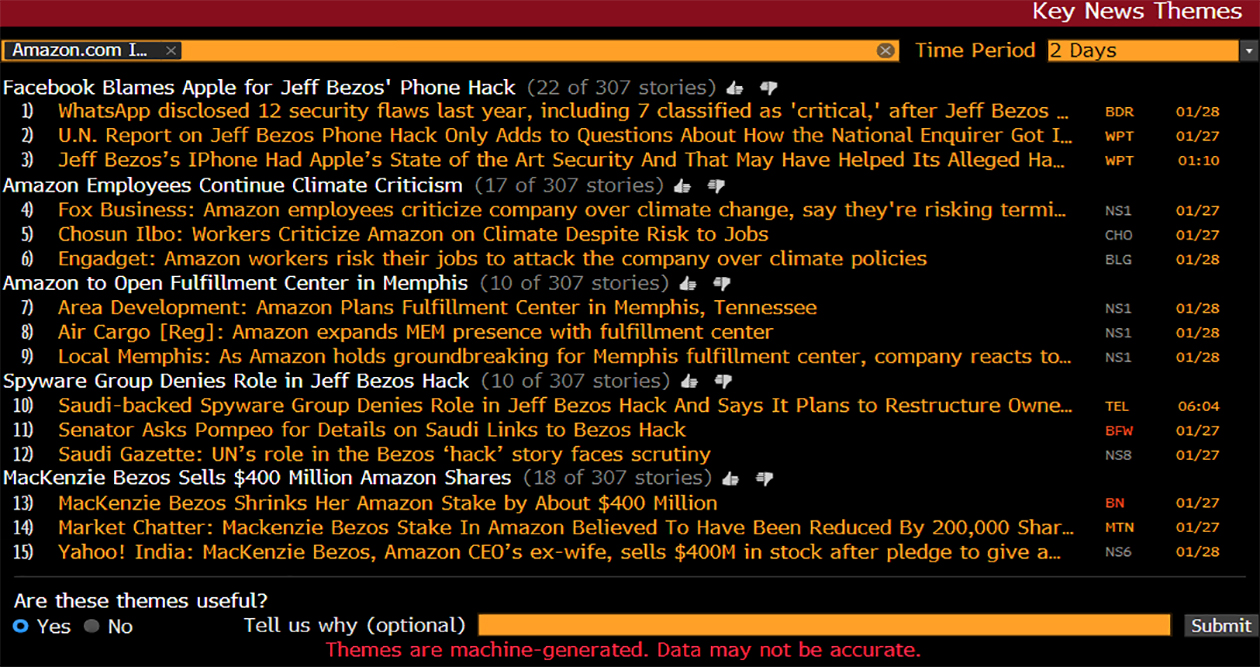}
      \caption{Screenshot (taken on 29 January 2020) of a query-driven interface for \nstm{} showing the overview for the company `Amazon.com'.}
      \label{fig:screenshot-company}
\end{figure*}

\begin{figure*}[!ht]
      \centering
      \includegraphics[width=\textwidth]{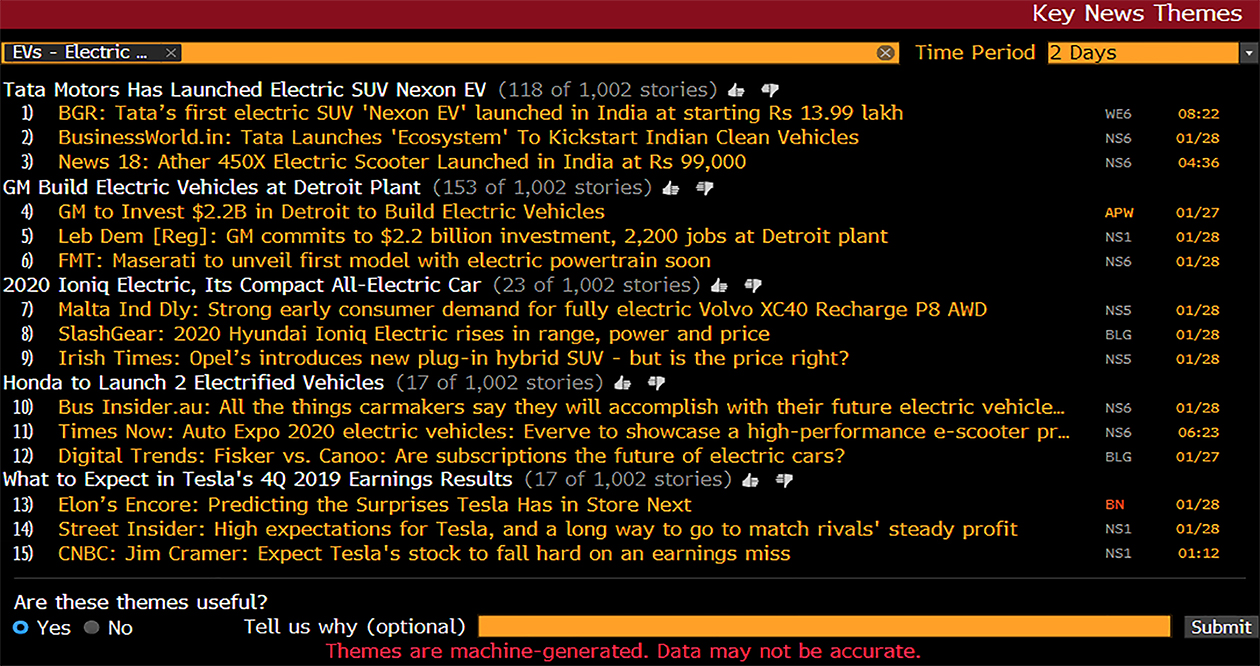}
      \caption{Screenshot (taken on 29 January 2020) of a query-driven interface for \nstm{} showing the overview for the topic `Electric Vehicles'.}
      \label{fig:screenshot-topic}
\end{figure*}

\begin{figure*}[!ht]
      \centering
      \includegraphics[width=\textwidth]{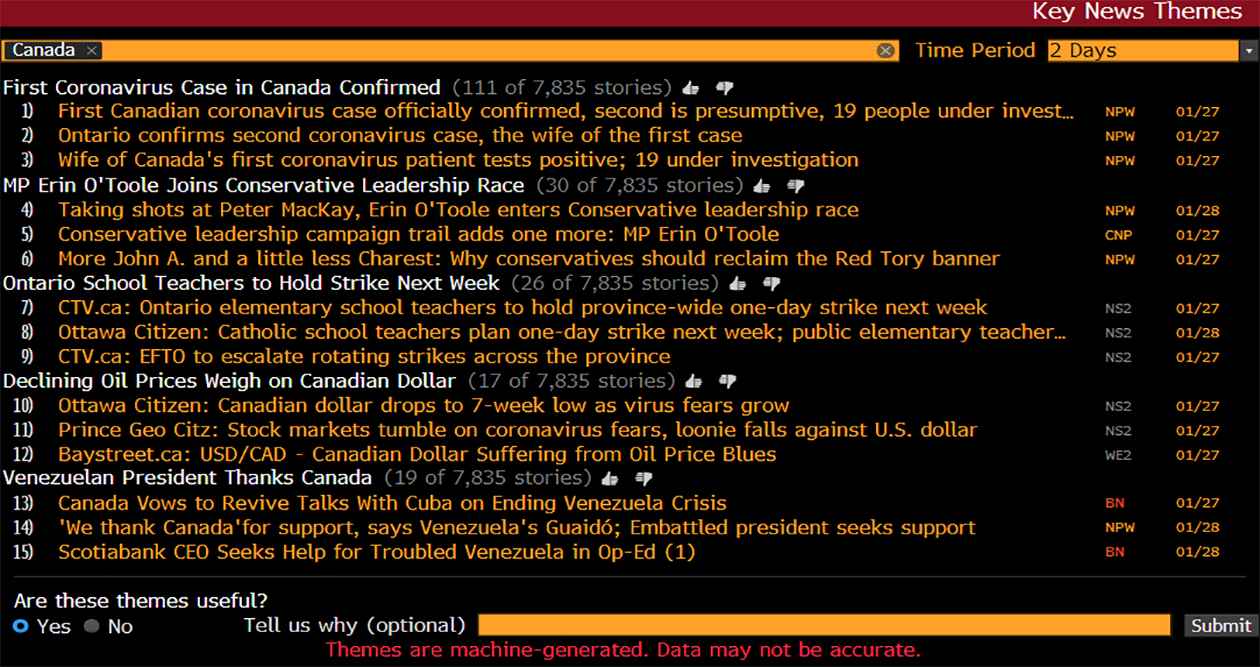}
      \caption{Screenshot (taken on 29 January 2020) of a query-driven interface for \nstm{} showing the overview for the region `Canada'.}
      \label{fig:screenshot-region}
\end{figure*}

\begin{figure*}[!ht]
      \centering
      \includegraphics[width=\textwidth]{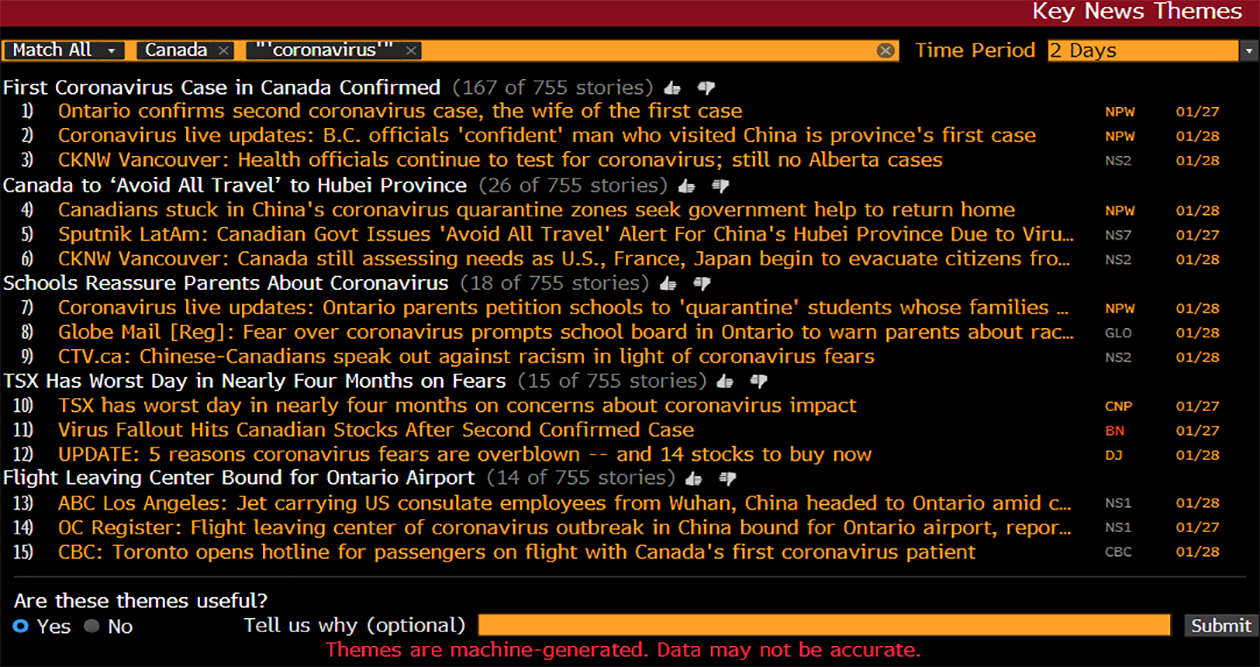}
      \caption{Screenshot (taken on 29 January 2020) of a query-driven interface for \nstm{} showing the overview for a complex query, including a keyword.}
      \label{fig:screenshot-complex-query}
\end{figure*}

\clearpage

\section{Screenshots of A Context-Driven User Interface}
\label{appendix:tren-screenshots}

\begin{figure*}[h!]
      \centering
      \includegraphics[]{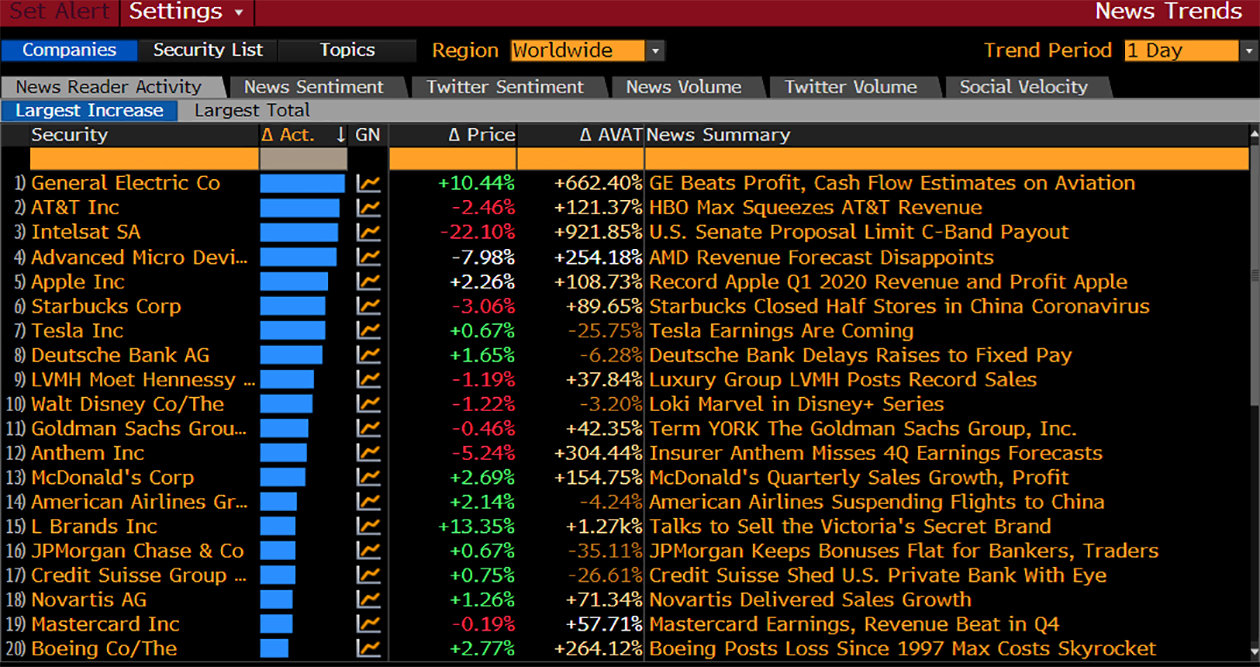}
      \caption{Screenshot (taken on 29 January 2020) of a context-driven application of \nstm{}. In the `Security' column are the companies that have seen the largest increase in news readership over the last day. Each entry in the `News Summary' column is the summary of the top theme provided by \nstm{} for the adjacent company.}
      \label{fig:screenshot-tren1}
\end{figure*}

\begin{figure*}[h!]
      \centering
      \includegraphics[]{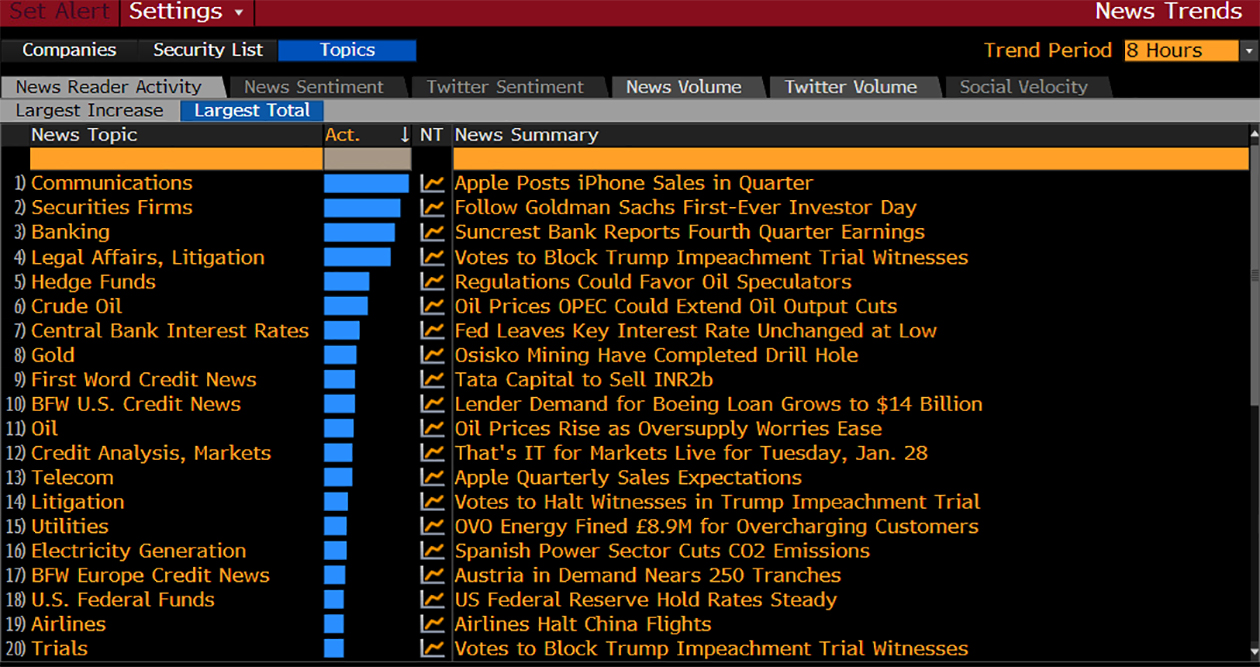}
      \caption{Screenshot (taken on 29 January 2020) of a context-driven application of \nstm{}. In the `News Topic' column are the topics that have seen the largest volume of news readership over the past 8 hours. Each entry in the `News Summary' column is the summary of the top theme provided by \nstm{} for the adjacent topic.}
      \label{fig:screenshot-tren2}
\end{figure*}

\end{document}


\section{Summary Candidate Ranking}

A large majority ($86\%$) of the candidates were found to be \emph{Terrible}, highlighting the necessity of additional ranking. We also found humans to be fairly consistent in terms of pairwise preferences in this task -- of the $3,033$ candidates which were labelled by multiple annotators, there were only $76$ conflicts (between any two annotators). Inter-annotator consistency was lowest when distinguishing between \emph{Great} and  \emph{Acceptable} candidates.